\documentclass[a4paper,fleqn]{cas-dc}

\usepackage[authoryear]{natbib}
\setcitestyle{authoryear,maxcitenames=2}

\def\tsc#1{\csdef{#1}{\textsc{\lowercase{#1}}\xspace}}
\tsc{WGM}
\tsc{QE}
\begin{document}
\let\WriteBookmarks\relax
\def\floatpagepagefraction{1}
\def\textpagefraction{.001}

\shorttitle{A protocol for evaluating robustness to H\&E staining variation in computational pathology models}    

\shortauthors{Schoenpflug et al.}  

\title [mode = title]{A protocol for evaluating robustness to H\&E staining variation in computational pathology models}  

% Author
\author[1,2]{Lydia A. Schoenpflug}[type=author,orcid=0009-0009-6703-9368]
\credit{Conceptualization, Data curation, Formal analysis, Investigation, Methodology, Project administration, Software, Validation, Visualization, Writing - Original Draft, Writing - Review \& Editing}

\author[3]{Nikki van den Berg}[type=author,orcid=0009-0003-2567-5311]
\credit{Conceptualization, Writing - Review \& Editing}

\author[1,2]{Sonali Andani}[type=author,orcid=0000-0003-3251-5108]
\credit{Conceptualization, Writing - Review \& Editing}

\author[4]{Nanda Horeweg}[type=author,orcid=0000-0002-8581-4753]
\credit{Conceptualization, Supervision, Writing - Review \& Editing}

\author[3]{Jurriaan Barkey Wolf}[type=author,orcid=0000-0002-7811-0280]
\credit{Software, Writing - Review \& Editing}

\author[3]{Tjalling Bosse}[type=author,orcid=0000-0002-6881-8437]
\credit{Conceptualization, Writing - Review \& Editing}

\author[1,2]{Viktor H. Koelzer}[type=author,orcid=0000-0001-9206-4885]
\credit{Conceptualization, Funding acquisition, Supervision, Writing - Review \& Editing}
\cormark[1]
\fnmark[1]
\ead{viktor.koelzer@usb.ch}

\author[1,2]{Maxime W. Lafarge}[type=author,orcid=0000-0001-9235-783X]
\credit{Conceptualization, Methodology, Project administration, Software, Supervision, Writing - Review \& Editing}
\fnmark[1]

\cortext[1]{Corresponding author}
\fntext[1]{These authors contributed equally as senior authors.}

\affiliation[1]{organization={Department of Biomedical Engineering, University of Basel},
            city={Allschwil},
            country={Switzerland}}
\affiliation[2]{organization={Institute of Medical Genetics and Pathology, University Hospital Basel},
            city={Basel},
            country={Switzerland}}
\affiliation[3]{organization={Department of Pathology, Leiden University Medical Center},
            city={Leiden},
            country={The Netherlands}}
\affiliation[4]{organization={Department of Radiation Oncology, Leiden University Medical Center},
            city={Leiden},
            country={The Netherlands}}
            
\begin{abstract}
Sensitivity to staining variation remains a major barrier to deploying computational pathology (CPath) models as hematoxylin and eosin (H\&E) staining varies across laboratories, requiring systematic assessment of how this variability affects model prediction. In this work, we developed a three-step protocol for evaluating robustness to H\&E staining variation in CPath models. Step 1: Select reference staining conditions, Step 2: Characterize test set staining properties, Step 3: Apply CPath model(s) under simulated reference staining conditions. Here, we first created a new reference staining library based on the PLISM dataset. As an exemplary use case, we applied the protocol to assess the robustness properties of 306 microsatellite instability (MSI) classification models on the unseen SurGen colorectal cancer dataset (n=738), including 300 attention-based multiple instance learning models trained on the TCGA-COAD/READ datasets across three feature extractors (UNI2-h, H-Optimus-1, Virchow2), alongside six public MSI classification models. Classification performance was measured as AUC, and robustness as the min-max AUC range across four simulated staining conditions (low/high H\&E intensity, low/high H\&E color similarity). Across models and staining conditions, classification performance ranged from AUC 0.769-0.911 ($\Delta=0.142$). Robustness ranged from 0.007-0.079 ($\Delta=0.072$), and showed a weak inverse correlation with classification performance (Pearson r=-0.22, 95\% CI [-0.34, -0.11]). %Comparing AUCs across simulated staining conditions, no condition universally outperformed other conditions.
Thus, we show that the proposed evaluation protocol enables robustness-informed CPath model selection and provides insight into performance shifts across H\&E staining conditions, supporting the identification of operational ranges for reliable model deployment. The reference library and code are publicly available at \url{https://huggingface.co/datasets/CTPLab-DBE-UniBas/staining-robustness-evaluation} and \url{https://github.com/CTPLab/staining-robustness-evaluation}.
\end{abstract}

\begin{graphicalabstract}
\includegraphics[width=\textwidth]{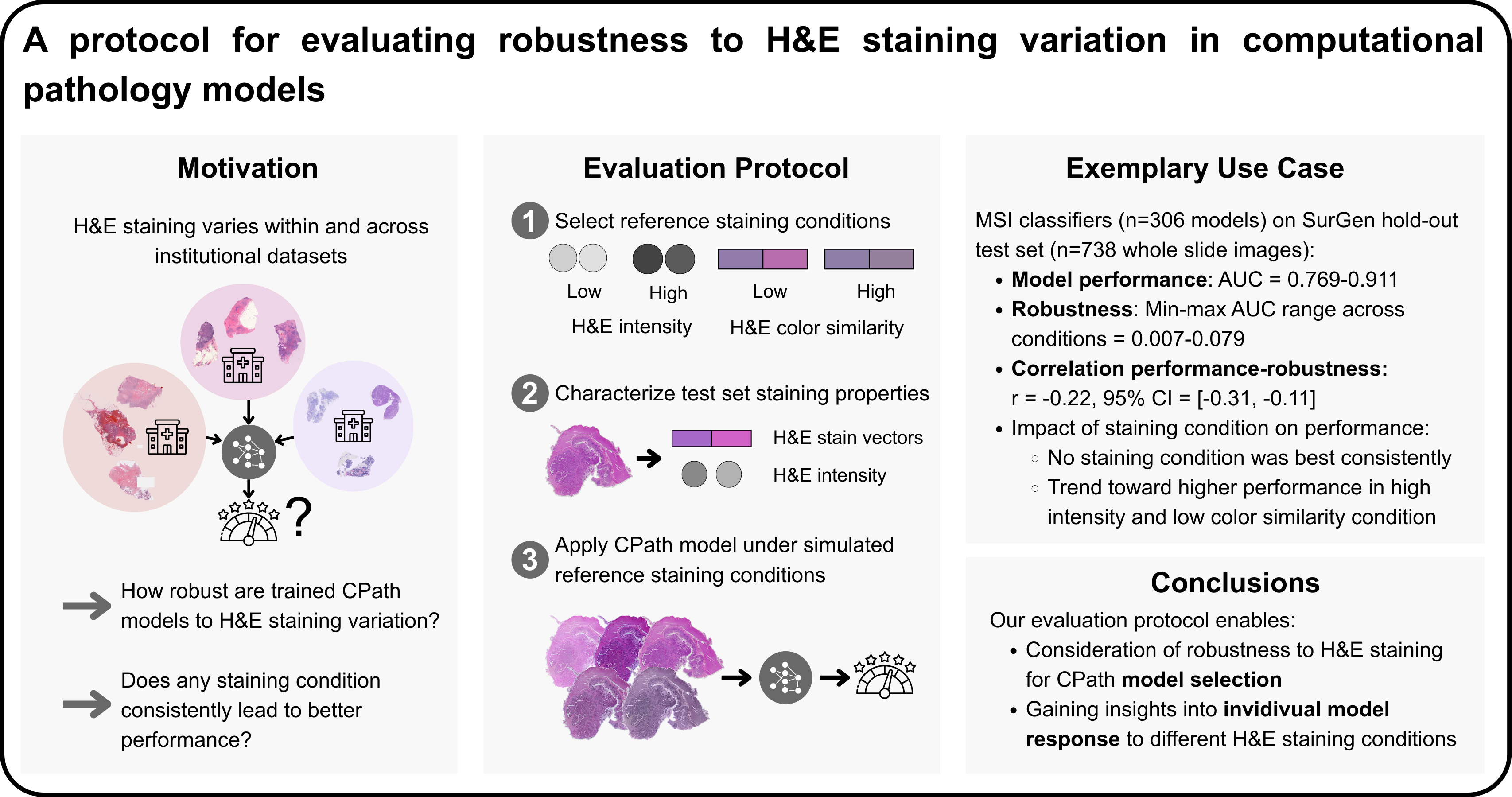}
\end{graphicalabstract}

\begin{highlights}
\item Evaluation protocol quantifies CPath model robustness to H\&E staining variation
\item Built a reference library of H\&E stain properties from the PLISM dataset
\item 306 CRC MSI classification models evaluated under four H\&E staining conditions
\item Robustness to staining variation is a key criterion for CPath model selection
\end{highlights}

%\nocite{*}

% Keywords

\maketitle

% Main text
\section{Introduction}\label{sec:Introduction}
Hematoxylin and eosin (H\&E) staining is central to diagnostic histopathology and forms the input of most computational pathology (CPath) pipelines. However, the appearance of Whole Slide Images (WSIs) of H\&E-stained specimens exhibits substantial variability within and across laboratories due to differences in staining protocols, reagent concentrations, batch effects, scanner characteristics, and tissue processing workflows \cite{Dunn2025}. These differences produce shifts in stain intensity, color, and can alter morphological information, which can affect model performance when CPath models are applied to WSIs of specimens prepared at different timepoints or institutions \cite{Folmsbee2019,Howard2021,Lin2025,Stacke2021,Tellez2019}. 
\begin{figure*}
    \centering
    \includegraphics[width=\textwidth,keepaspectratio]{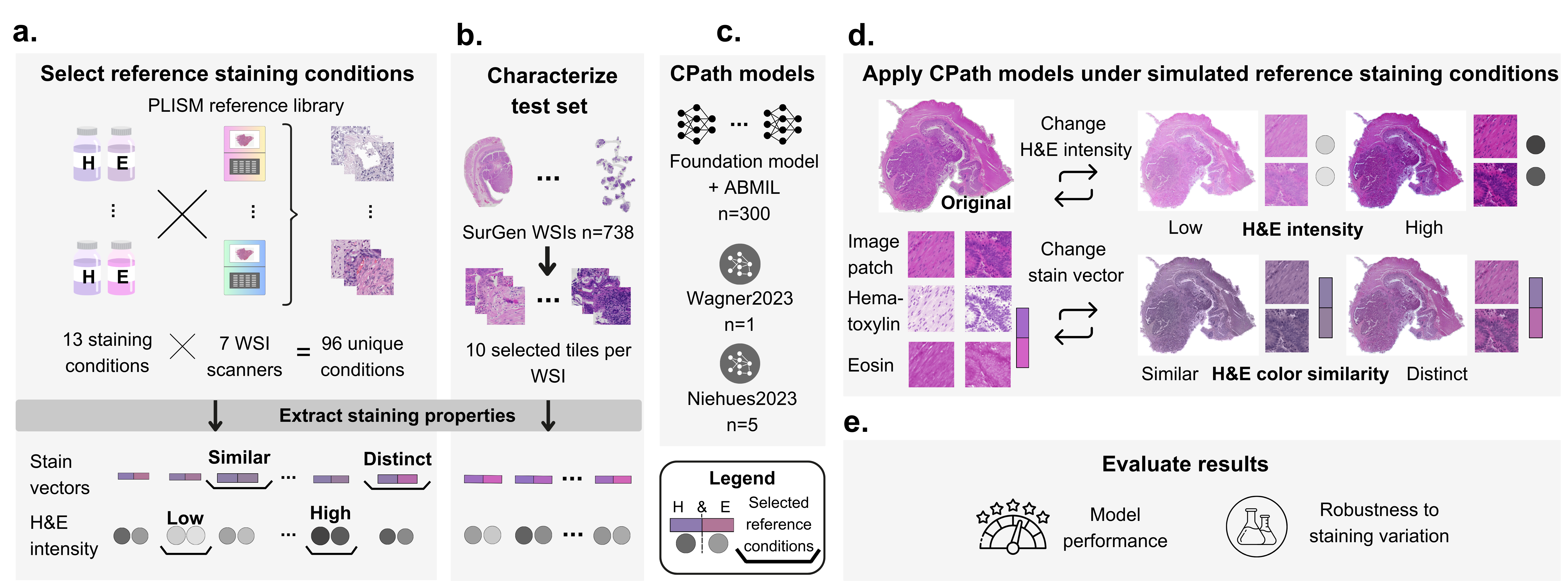}
    \caption{A protocol for evaluating robustness to staining variation of CPath models. a) Select reference staining conditions based on the reference libray created from the PLISM dataset. b)  Characterize SurGen test set staining properties. c) State-of-the art MSI classification models, with n=300 models reflecting plausible state-of-the-art models which were trained on TCGA COADREAD using one of three foundation models as feature extractors (Uni2-h, HOptimus1, Virchow2) and ABMIL for aggregation. We also considered n=6 publicly available models. d) Infer MSI classification models under four simulated reference staining conditions. e) Evaluation of results by measuring model performance and robustness to staining variation.
    }
    \label{fig:fig1-protocol-overview}
\end{figure*}

Prior work addressing staining-related domain shift can be broadly grouped into development-focused and evaluation-focused approaches. Development-focused methods, including stain normalization \cite{Macenko2009,Janowczyk2017, Sim2025, Du2025}, stain augmentation \cite{Balkenhol2018,Tellez2019,Bentaieb2018,Nishar2020,Shen2022,Marini2023}, and domain-adversarial training \cite{Lafarge2017,Otalora2019}, all aim to improve robustness during training. However, a critical gap arises in the context of modern computational pathology pipelines: existing methods primarily focus on improving encoder-level training, assuming encoders are trained locally. At the same time the field has shifted toward large pre-trained foundation models, whose training is restricted to institutions with access to massive datasets and computational resources (e.g. UNI: >100k WSIs, 32 GPUs) \cite{Xiong2025}. As a result, most researchers no longer modify the encoder but instead train lightweight attention-based multiple instance learning (ABMIL) classifiers on top of features extracted by frozen foundation models, further referred to as two-stage training pipelines. In this setting, training-time strategies such as stain augmentation or normalization have reduced influence \cite{Woelflein2023}, as their impact is constrained by the pre-trained feature representations. At the same time, current foundation models are not fully invariant to domain shifts, including staining variability: extracted features retain institution-specific information \cite{Lin2025-2,Jong2025,Koemen2024,Koemen2025}. Yet, pipelines based on foundation models generally outperform previous approaches, suggesting that adequate pretraining confers a measurable improvement in robustness to staining variation and reduces, but does not eliminate, sensitivity to H\&E variability.  Evaluation-focused studies have demonstrated that staining variability can affect performance more strongly than many common artifacts \cite{Wang2021,Foote2022,Schoemig-Markiefka2021,Vu2022}. Yet, existing evaluation strategies often rely on image-based references \cite{Schoemig-Markiefka2021,Vu2022}, generative transformations \cite{Springenberg2023}, fixed multiplier perturbations \cite{Wang2021}, or re-staining/re-scanning protocols \cite{Ruiz2023,Sun2023,Chai2026}, which make it difficult to attribute performance changes to specific, quantifiable staining properties or to situate perturbations within a defined staining reference space. Taken together, these observations highlight a need for systematic evaluation methods that quantify the robustness to staining variation of trained two-stage CPath pipelines at inference time, while anchoring perturbations to defined staining properties. Such methods can inform both model selection and diagnostic slide preparation, supporting data-driven quality control. 

To address this deficit, we introduce a protocol for systematically evaluating the robustness of computational pathology models under realistic H\&E staining variations (Figure \ref{fig:fig1-protocol-overview}). Our approach simulates plausible staining conditions while preserving tissue structure, thereby enabling quantitative assessment of model robustness. We demonstrate this protocol using microsatellite instability (MSI) classification in colorectal cancer, evaluating multiple state-of-the-art models and providing a methodological basis for evidence-based slide preparation and quality control in clinical workflows.

\section{Methods}\label{sec:Methods}
\subsection{Stain decomposition and controlled recomposition framework}\label{subsec:stain-decomposition}
We modeled the RGB pixel intensities of a histology image $I \in [0,255]^{H \times W \times 3}$ as a function of the optical density (OD) using the Beer-Lambert law
\begin{equation}
\mathrm{OD} = -\log\!\left(\frac{I}{I_0}\right),
\end{equation}
where $I_0$ denotes the illumination constant (typically $255$). 
Following \cite{Macenko2009}, the OD of each pixel is assumed 
to lie in the subspace spanned by the hematoxylin and eosin stain vectors $\mathbf{s}_H, \mathbf{s}_E \in \mathbb{R}^3$. We further include a third residual vector $\mathbf{s}_R$ orthogonal to the H\&E plane to account for residual variation. Using this framework, each pixel's optical density is represented as
\begin{equation}
\mathrm{OD} \approx i_H \mathbf{s}_H + i_E \mathbf{s}_E + i_R \mathbf{s}_R ,
\end{equation}
where $i_H, i_E, i_R \ge 0$ denote the stain intensities, quantifying the per-pixel staining magnitude.
For image-level quantification, the H\&E stain intensity is defined as the 95th percentile across all pixels. To robustly identify the H\&E stain vectors, the OD matrix of filtered pixels is analyzed via singular value decomposition. The two dominant singular vectors define the H\&E plane, in which the stain vectors are subsequently identified, and the third singular vector defines the residual vector. Pixels with extreme OD values are excluded to avoid artifacts. For staining simulations, OD projections are scaled to target H\&E intensities and recombined using the target stain vectors, with the residual component reduced by a factor of $0.01$ to minimize its effect.

\subsection{Datasets}\label{subsec:datasets} 
\subsubsection{PLISM}\label{subsubsec:PLISM} 
The Pathology Images of Scanners and Mobilephones (PLISM) dataset \cite{Ochi2024-PLISM} comprises aligned image patches from 46 human tissue types, stained under 13 H\&E staining conditions and digitized using 13 imaging devices, with recorded  slide preparation parameters\footnote{\url{https://huggingface.co/datasets/owkin/plism-dataset-tiles}, last accessed 04.12.2025.}. This design enables extraction of realistic H\&E staining intensities and stain vectors for downstream simulation. As this study is focused on the analysis of WSIs, we restricted the analysis to the subset of images patches of the PLISM dataset captured with WSI scanners (PLISM‑wsi). The final subset includes 310,947 patches, corresponding to 3,417 aligned patches across 91 unique stain-device combinations (Appendix Table A.1 and A.2). We excluded the HR and GIV staining conditions for our study, as we consider the overnight hematoxylin exposure to produce an unrealistically intense staining.

\subsubsection{TCGA COAD/READ}\label{subsubsec:TCGA} 
The TCGA Colon and Rectum Adenocarcinoma cohorts (COAD, READ) were used as the development set for training and validating a set of plausible “black-box” MSI classifiers. A total of 625 diagnostic WSIs of H\&E-stained formalin-fixed paraffin-embedded (FFPE) resection specimens from 616 patients were downloaded from the Genomic Data Commons portal\footnote{\url{https://portal.gdc.cancer.gov/}, last accessed 04.12.2025.} of which 25 WSIs were excluded due to file corruption, yielding 600 WSIs from 591 patients. MSI status was obtained for all patients from two sources available through cBioPortal\footnote{\url{ https://www.cbioportal.org/}, last accessed 04.12.2025.}: (i) the PanCancer Atlas \cite{Bonneville2017-MSI1}, providing next-generation sequencing (NGS)-derived MSI calls using MANTIS \cite{Kautto2016-MANTIS}, and (ii) the original TCGA publications \cite{TCGA2012-TCGACOADREAD} which reported PCR-based MSI status. When both labels were available, NGS-derived MSI calls were prioritized to ensure a consistent, genome-wide reference standard across the cohort. 

\subsubsection{SurGen}\label{subsubsec:SurGen} 
The SurGen dataset \cite{Myles2025-SurGen} is a publicly available\footnote{\url{ https://www.ebi.ac.uk/biostudies/bioimages/studies/S-BIAD1285}, last accessed 04.12.2025.} collection of WSIs of FFPE H\&E-stained CRC resection and biopsy specimens and was used solely as a holdout test set in this study. It comprises 1,020 WSIs from 843 patients and is divided into two subsets: SR386, containing primary colorectal tumors with matched genomic and clinical annotations, and SR1482, which includes both primary and metastatic cases. MSI status was derived from either PCR-based fragment analysis or mismatch-repair (MMR) immunohistochemistry (IHC), with PCR-based calls preferred where available. For IHC-derived labels, loss of expression of one or more MMR proteins was interpreted as microsatellite instable, intact expression of all MMR proteins was interpreted as microsatellite stable. Patients with WSIs exclusively from non-colorectal metastatic sites (n=92), missing MSI status (n=11), or providing insufficient tissue (n=2) were excluded, resulting in 738 patients with 892 WSIs. When multiple slides were available for the same patient, the first available WSI was selected.

\subsection{Models}\label{subsec:models}
\subsubsection{Simulated models}\label{subsec:simulated-models}
To reflect the diversity of models that may realistically be developed or received by clinical implementers, we simulated a set of ABMIL models trained on TCGA COAD/READ using foundation-model-based pipelines. Tissue detection was performed on all WSIs using the tissue segmentation model of \cite{Bandi2019-TissueSgm}, followed by tiling into $224\times224$ px patches at 1µm/px for feature extraction. No stain augmentation or normalization was applied, following established ABMIL training frameworks \cite{Woelflein2023}.
To induce realistic variability, we sampled across multiple hyperparameters: three foundation model feature extractors (UNI2-h\footnote{\url{https://huggingface.co/MahmoodLab/UNI2-h}, last accessed 04.12.2025.} \cite{Chen2024-UNI}, H-Optimus-1\footnote{\url{https://huggingface.co/bioptimus/H-Optimus-1}, last accessed 04.12.2025.} \cite{Hoptimus1}, Virchow2\footnote{\url{https://huggingface.co/paige-ai/Virchow2}, last accessed 04.12.2025.} \cite{Zimmermann2024-Virchow2}) with 100 models per extractor, arbitrary random initialization seeds, two weight decay values ($0$ or $10^{-4}$), MSI-stratified 70:30 patient-level train-validation splits with or without institutional overlap, and random exclusion of 0–10 institutions per model while retaining at least 90\% of the development dataset. This approach simulates variability in data splits, training methodology, and institutional coverage.
Slide-level aggregation of patch-level features was performed using a gated ABMIL model adapted from \cite{Lu2021}, with an additional pre-compression fully connected layer. Models were trained using AdamW \cite{Loshchilov2019-AdamW} with learning rate $5\cdot10^{-5}$, momentum 0.95/0.999, cosine annealing, batch size 1, early stopping with five-epoch patience based on validation AUC, and a maximum of 30 epochs.

\subsubsection{Public models}\label{subsec:public-models}
In addition to the simulated models, we included six publicly available, trained MSI classification models from two publications. The first, Wagner2023 \cite{Wagner2023}, uses the CTransPath \cite{Wang2022-CTransPath} model for feature extraction from $512 \times 512$px image patches at 0.5µm/px (resized to $224 \times 224$px, 1.14µm/px), followed by a transformer-based MIL (TransMIL) aggregator. The TransMIL model was trained on WSIs from over 13,000 CRC patients across 16 cohorts. The five other models, Niehues2023 \cite{Niehues2023}, use the RetCCL \cite{Wang2023-RetCCL} model for feature extraction with identical input format as Wagner2023 ($224 \times 224$px, 1.14µm/px) and gated ABMIL aggregation. These five models were trained with different dataset splits under a 5-fold cross-validation framework. Feature extractors and pretrained MIL weights were obtained from the authors’ publicly released repositories\footnote{%
CTransPath: \url{https://huggingface.co/jamesdolezal/CTransPath}; 
RetCCL: \url{https://github.com/Xiyue-Wang/RetCCL}; 
Wagner2023: \url{https://github.com/peng-lab/HistoBistro/tree/main/CancerCellCRCTransformer}; 
Niehues2023: \url{https://github.com/KatherLab/crc-models-2022/tree/main/Quasar_models/Wang\%2BattMIL/isMSIH};
all last accessed 15.01.2026.}. Inference was performed without stain normalization and consisted of feature extraction using CTransPath or RetCCL, followed by aggregation with the respective pretrained MIL models to obtain slide-level MSI classifications.

\subsection{Protocol for evaluating robustness to H\&E staining variation}\label{subsec:protocol}
Our protocol for evaluation of CPath model robustness to H\&E staining variation comprises three stages. First, reference staining conditions are selected based on a reference library. Here, we first created a new reference library from the PLISM dataset. Second, the test set staining properties are characterized, specifically slide-wise stain vectors and intensities are extracted. Third, inference of the CPath model is performed under a controlled set of simulated staining conditions. Here, we present the application of this protocol to MSI classification models, evaluated and compared using the holdout SurGen test set.

\subsubsection{Step 1: Select reference staining conditions}\label{subsubsec:protocol-step1}
To enable reference selection we first created a new reference library from the PLISM dataset. Only tiles meeting quality criteria (Appendix Table B.1) were included. Stain vectors and intensities were extracted for these tiles as described in section \ref{subsec:stain-decomposition}, and median values across tiles were computed for each stain-device combination to define representative H\&E stain vectors and intensities. Stain intensity was quantified as the 95th percentile of hematoxylin and eosin intensities in the stain-separated projections, capturing well-stained tissue while avoiding extreme artifacts (e.g., blood). For simulating intensity-based changes, the conditions with lowest and highest H\&E intensity were selected. To simulate changes of H\&E stain colors, we selected stain vector references reflecting common diagnostic hematoxylin formulations: Harris for the most distinct and Gill for the most similar H\&E stain colors, where similarity was measured as the angle between H\&E vectors in OD space. These references enable realistic simulation of both intensity- and color-based variations in H\&E staining. Exemplary tiles and staining details for each reference can be found in Appendix Figure B.1 and Table B.2. 
All the stain vectors and intensities derived from the PLISM dataset were released as a public reference library \footnote{PLISM stain references on HuggingFace: \url{https://huggingface.co/datasets/CTPLab-DBE-UniBas/staining-robustness-evaluation/tree/main/plism-wsi_stain_references}}.

\subsubsection{Step 2: Characterize test set staining properties}\label{subsubsec:protocol-step2}
For each slide of the SurGen dataset, slide-level stain vectors and H\&E intensities were obtained by sampling 10 tiles (448px $\times$ 448px at 0.5µm/px) per slide that passed the tile quality criteria (Appendix Table B.1). In rare cases where artifact colors affected decomposition (e.g. pink, orange, brown, or dark-red blood), quality criteria were manually adjusted until visual review of the derived stain vectors showed alignment with WSI stain appearance. Stain vectors and H\&E intensity values were extracted as described for the PLISM reference conditions, and slide-level representatives were computed as the median across the sampled tiles. Slide-wise stain intensities and vectors then served as the basis for staining simulations during inference. 
All the slide-level stain vectors and intensities were made publicly available\footnote{SurGen stain references on HuggingFace: \url{https://huggingface.co/datasets/CTPLab-DBE-UniBas/staining-robustness-evaluation/tree/main/surgen_stain_properties}}.

\subsubsection{Step 3: Apply CPath model under simulated reference staining conditions}\label{subsubsec:protocol-step3}
Inference under controlled, simulated staining conditions on SurGen WSIs consisted of tissue detection with the tissue segmentation model of \cite{Bandi2019-TissueSgm}, followed by tiling into patches, matching the input size and resolution of the respective feature extractor as specified in section \ref{subsec:simulated-models} and \ref{subsec:public-models}. Tiles are then decomposed into hematoxylin and eosin channels using the slides' stain vectors, and are further recomposed based on the target PLISM references to simulate the target staining conditions as described in section \ref{subsec:stain-decomposition}. Experiments covered four PLISM-defined conditions: low vs. high H\&E intensity and high vs. low H\&E color similarity. In the intensity conditions, only stain intensities were varied (stain vectors fixed), whereas in the color similarity conditions only stain vectors were varied (intensities fixed). Features were extracted once for all foundation models and conditions. 

\subsection{Evaluation metrics}\label{subsec:evaluation-metrics}
Model performance was quantified as the AUC on the reference staining condition, and robustness to staining variation as the min-max range of AUCs across the five staining conditions (reference and four simulated staining conditions). Models were considered high-performing with a reference AUC over 90\% and highly robust with a min-max AUC range of less than 3\%. Confidence intervals were estimated using 1,000 bootstrap resamples of the SurGen cohort. For each iteration, AUCs were computed for all five conditions, yielding a bootstrap-specific min-max range (robustness) and reference AUC (performance). To compare performance between the reference and simulated staining conditions, we calculated the difference in AUC ($\Delta$AUC). Distribution and variability of $\Delta$AUCs were summarized using median, quantile and minimum statistics. Further, staining conditions were ranked by AUC for each model, and we recorded how many models achieved their highest AUC under each condition. For visualization purposes, staining characteristics were summarized using stain intensity, stain similarity as the angle between H\&E vectors in OD space, and stain hue quantified as the hue angle ($h^\circ = \arctan2(b^*, a^*)$) in CIELab space. 

\subsection{GPU Usage and Code availability}\label{subsec:gpu-usage}
All code was implemented in Python, model training and inference was done with PyTorch. Training and inference were performed on NVIDIA H200 GPUs, totaling 444 GPU hours. Energy usage was measured via nvidia-smi and totaled 208.45 kWh (Appendix Table C.1). 
Code for stain vector and intensity extraction, model training, inference, analysis, and visualization is publicly available on GitHub\footnote{GitHub Repository: \url{https://github.com/CTPLab/staining-robustness-evaluation/tree/main}}. 
We further make PLISM and SurGen stain vectors and intensities, all 300 trained models and experiment results publicly available\footnote{HuggingFace Repository: \url{https://huggingface.co/datasets/CTPLab-DBE-UniBas/staining-robustness-evaluation/tree/main}}. 

\begin{figure*}
    \centering
    \includegraphics[width=\textwidth,keepaspectratio]{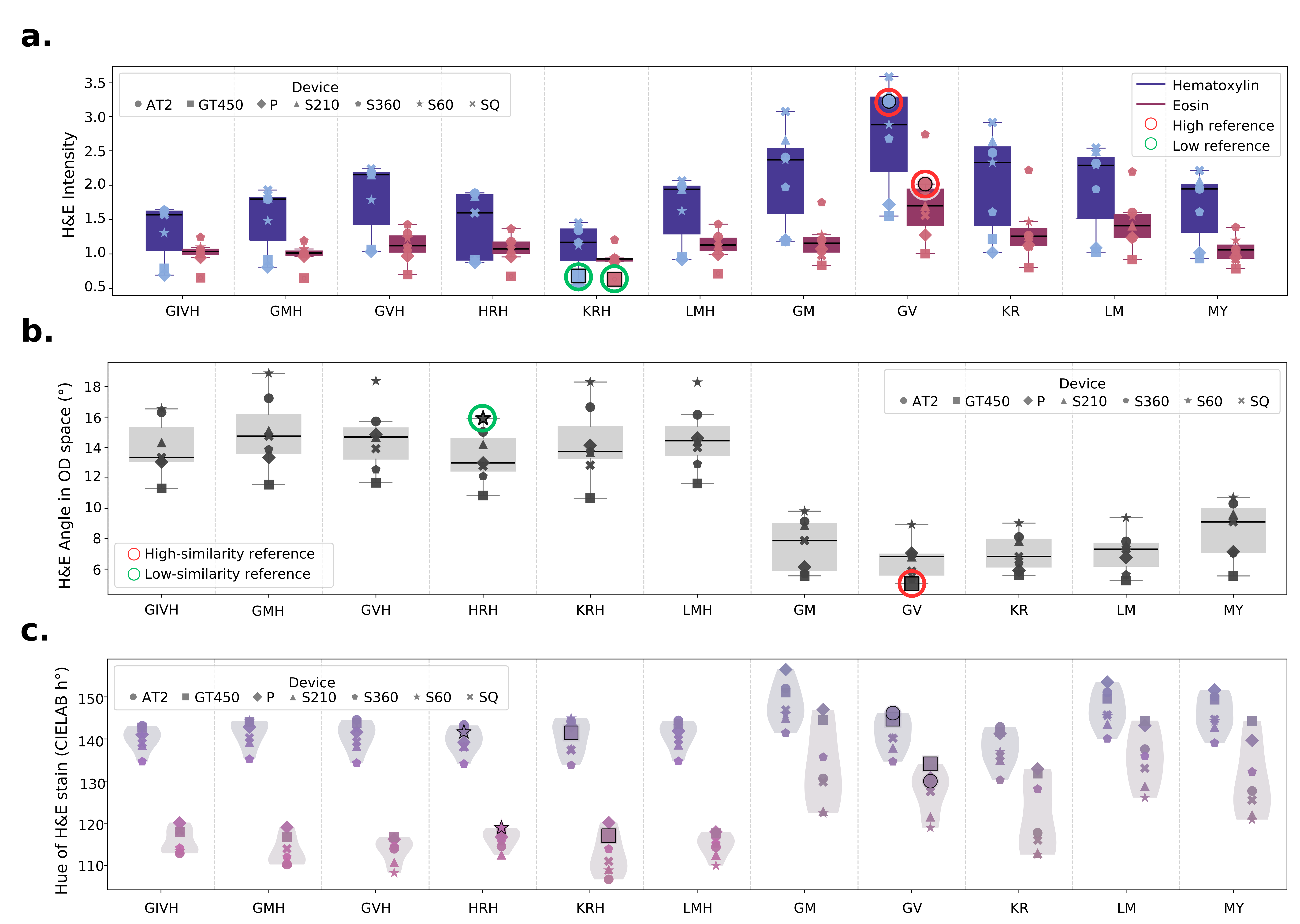}
    \caption{Staining characteristics of PLISM staining condition-device combinations. a) Intensity of Hematoxylin and Eosin, b) Angle between H\&E stain vector in OD space, c) Distribution of H\&E hues, measured as hue h° in CIELab space; left violin: Hematoxylin, right violin: Eosin. Marker colors correspond to RGB stain colors. The selected reference conditions (low and high intensity; low and high H\&E color similarity) are circled in red and green respectively and highlighted with a black frame. For staining condition and device abbreviations please refer to Appendix Table A.1 and A.2.
    }
    \label{fig:fig2-plism}
\end{figure*}

\section{Results}\label{sec:Results}
\subsection{PLISM-derived stain and intensity references}\label{subsec:Results-PLISM-stains}
To establish a benchmark for realistic H\&E staining changes, we first characterized the H\&E intensities and stain vectors for each unique PLISM stain-device condition. Consistent with the PLISM study design, conditions intended to be more distinct in H\&E color (ending with “H”) showed larger angles between H\&E stain vectors compared to similar H\&E color conditions (Figure \ref{fig:fig2-plism}a+b). This pattern was also reflected in H\&E hues, which displayed greater separation for distinct stain color conditions (Figure \ref{fig:fig2-plism}c). Furthermore, we note that PLISM conditions optimized for distinct H\&E colors exhibited lower H\&E intensities compared to higher similarity conditions (median H: 1.58 vs 2.10; median E: 1.04 vs 1.35). This difference did not map consistently onto a single recorded slide preparation parameter (e.g., hematoxylin formulation, exposure duration, or number of dehydration steps), suggesting a multifactorial relationship. Hematoxylin intensity varied more across staining conditions than across WSI scanners (mean std across stains = 0.56 vs. across devices = 0.47). In contrast, stain similarity and Eosin intensity was impacted more strongly by WSI scanning effects, with higher standard deviation across devices than across stainings (H\&E stain angle mean std across devices = 3.7° vs. across stainings = 1.9°; Eosin intensity mean std across stainings = 0.28 vs. across devices = 0.25).

\subsection{Distribution of H\&E staining intensity and colors in the SurGen dataset}\label{subsec:results-surgen}
\begin{figure}
    \centering
    \includegraphics[width=\columnwidth]{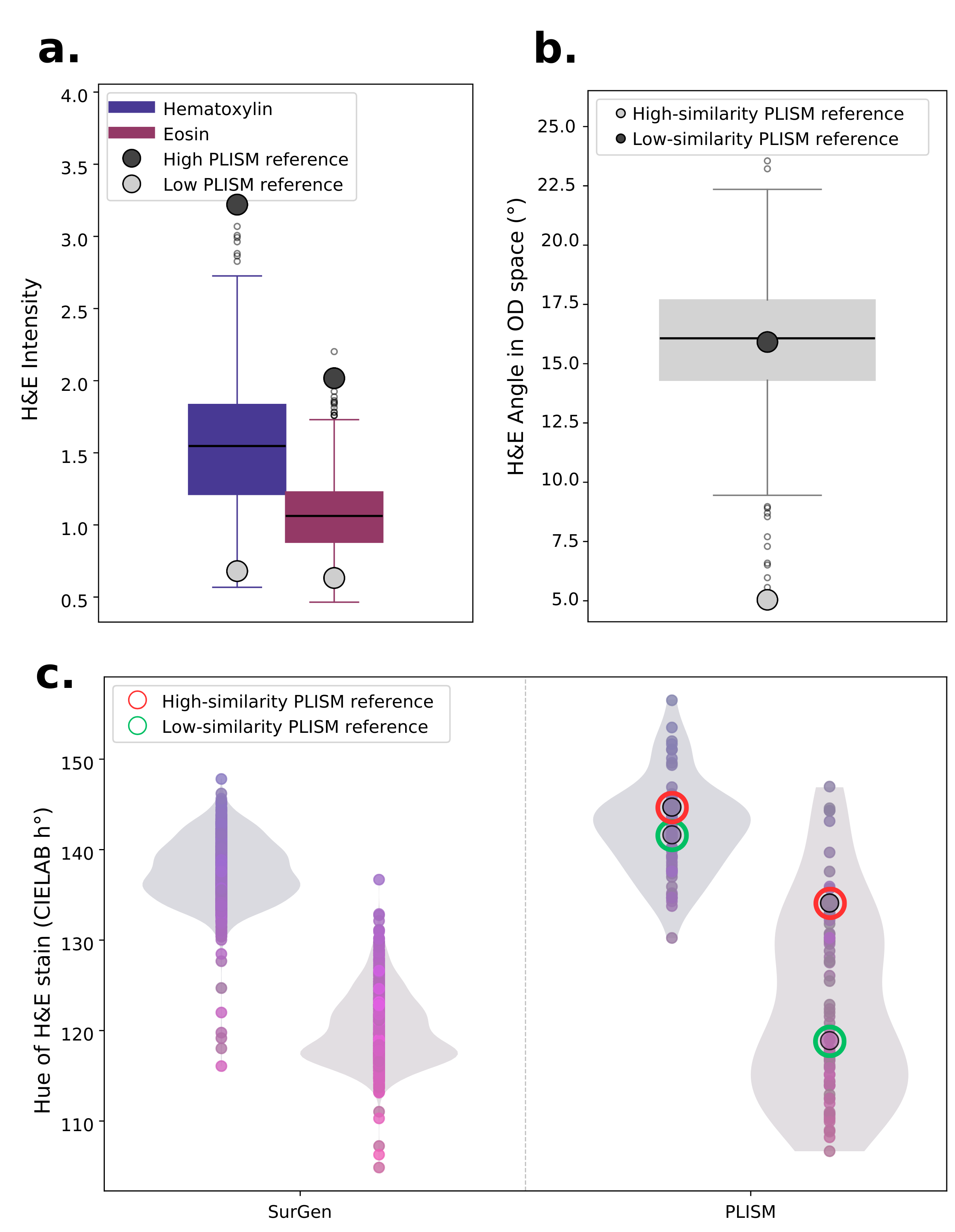}
    \caption{Staining characteristics of SurGen WSIs. a) Intensity of Hematoxylin and Eosin, b) Angle between H\&E stain vector in OD space, c) Distribution of H\&E hues, measured as hue h° in CIELab space; left violin: Hematoxylin, right violin: Eosin. Marker colors correspond to RGB stain colors; low and high color similarity PLISM references are circled in green and red respectively.}
    \label{fig:fig3-surgen}
\end{figure}

Analysis of SurGen slide-level H\&E intensities and stain vectors revealed substantial intrinsic variability, encompassing, and in some cases, exceeding the selected PLISM reference conditions (Figure \ref{fig:fig3-surgen}). Hematoxylin and eosin intensities ranged from 0.57-3.10 and 0.46-2.20, slightly extending beyond the low (H = 0.68, E = 0.63) and high (H = 3.22, E = 2.02) PLISM reference intensities. H\&E hues also marginally extended the PLISM reference range. H\&E stain color similarity, measured as the angle between H\&E vectors, spanned 5.1°-23.6°, covering the PLISM “high-similarity” reference (5.0°) and surpassing the “low-similarity” reference (15.9°). 

\subsection{Robustness-performance relationship in MSI classification}\label{subsec:results-robustness-performance}
\begin{figure*}
    \centering
    \includegraphics[width=\textwidth,keepaspectratio]{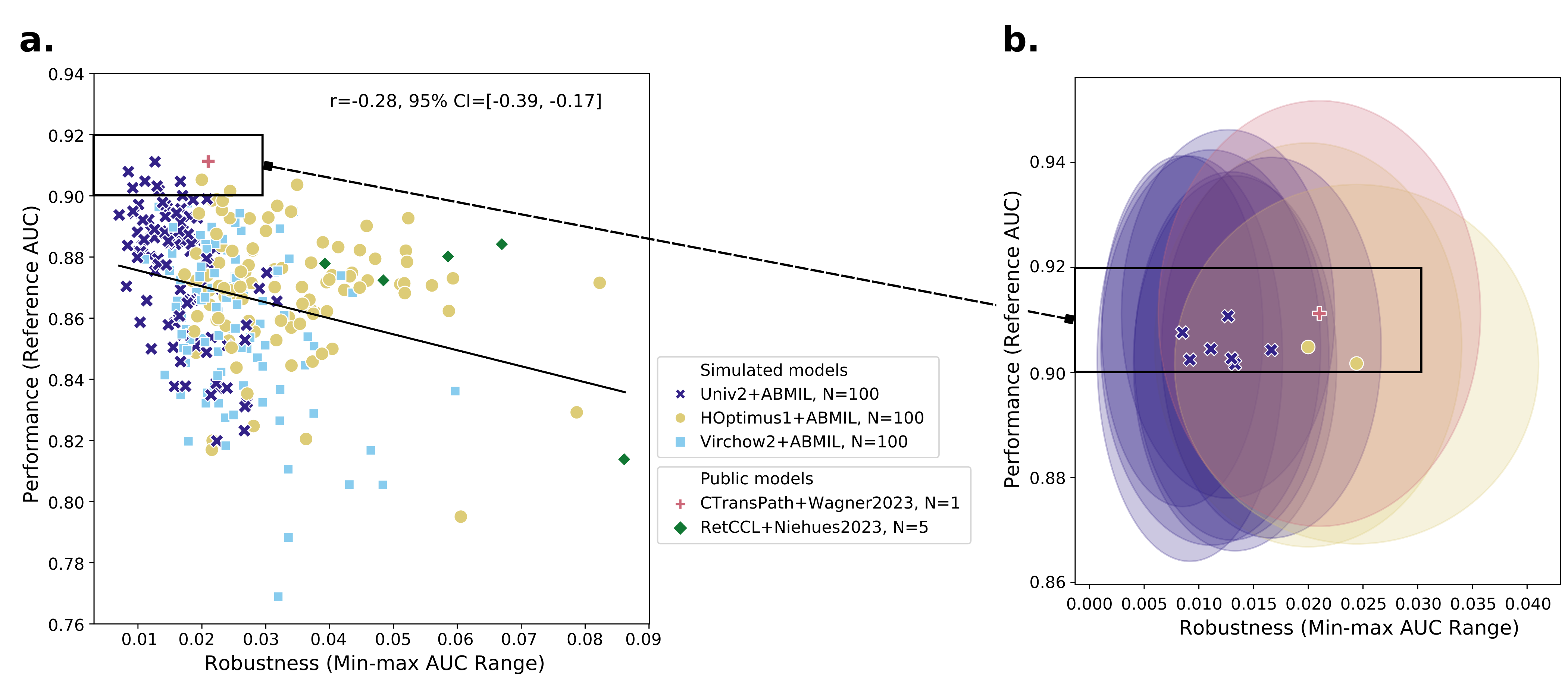}
    \caption{Performance-robustness relationship across MSI classification models: a) Performance (AUC of reference condition) versus robustness (min-max AUC across all stain conditions) for all evaluated models (n=306). We report Pearson correlation between performance and robustness with 95\% CIs. b) Top models (Performance > 0.90, Robustness <0.03; n=10). Dots indicate bootstrapped means (n=1000 iterations); ellipses represent 95\% CIs for both performance and robustness.
    }
    \label{fig:fig4-robustness-performance}
\end{figure*}

To examine how model performance relates to robustness under staining simulations, we evaluated model performance on the SurGen dataset under the original and four simulated H\&E staining conditions for the 306 MSI classification models. Across all models, reference performance ranged from AUC 0.769-0.911, robustness to staining variation ranged from 0.007-0.079 and exhibited a weak inverse association (r = -0.28, 95\% CI = [-0.39, -0.17], n = 306), indicating that higher reference AUC does not necessarily imply greater robustness to staining variability (Figure \ref{fig:fig4-robustness-performance}a). However, this varied depending on the foundation model. A moderate negative correlation was observed for UNI2-h+\allowbreak ABMIL (r = -0.51, 95\% CI = [-0.63, -0.39], n = 100) and Virchow2+\allowbreak ABMIL (r = -0.36, 95\% CI = [-0.52, -0.19], n=100), while no significant association was observed for H-Optimus-1+\allowbreak ABMIL (r = -0.14, 95\% CI = [-0.36, 0.10], n = 100) and the RetCCL+\allowbreak Niehues2023 models (r = -0.75, no 95\% CI due to low sample size, n = 5). Reference AUC distributions across foundation models are summarized in Appendix D, showing differences in median performance (median AUC UNI2-h: 0.881, H-Optimus-1: 0.865, Virchow2: 0.856, across N=100 models), but substantial overlap in interquartile ranges (UNI2-h: [0.860, 0.892], H-Optimus-1: [0.848, 0.878], Virchow2: [0.838, 0.873).
\begin{table*}
\centering
\caption{Effects of simulated H\&E staining conditions on all models ($n=306$) and top-performing, robust models ($n=10$). For each condition, the table reports the number of models for which it performed best, the median $\Delta$AUC relative to reference with 95\% CI, and the worst-case AUC decrease compared to reference.}
\label{tab:tab2-all-models-staining}
\begin{tabular}{lccc}
\textbf{All models (n=306)} & &  &  \\
\hline
\textbf{Staining Condition} & \textbf{Best model count} & \textbf{Median $\Delta$AUC [95\% CI]} & \textbf{Worst-case AUC decrease} \\
\hline
Reference & 65 & - & - \\
Low intensity & 30 & -0.50\% [-1.13\%, -0.05\%] & -4.50\% \\
High intensity & 127 & 0.12\% [-0.65\%, 0.73\%] & -4.36\% \\
Low H\&E color similarity & 51 & -0.07\% [-0.52\%, 0.34\%] & -3.17\% \\
High H\&E color similarity & 33 & -0.57\% [-1.57\%, 0.11\%] & -7.78\% \\
\hline
 &  &  &  \\
\end{tabular}

\begin{tabular}{lccc}
\textbf{Top models (n=10)} & &  &  \\
\hline
\textbf{Staining Condition} & \textbf{Best model count} & \textbf{Median $\Delta$AUC [95\% CI]} & \textbf{Worst-case AUC decrease} \\
\hline
Reference & 0 & - & - \\
Low intensity & 0 & -0.46\% [-0.63\%, -0.07\%] & -1.08\% \\
High intensity & 8 & 0.50\% [0.36\%, 0.73\%] & -0.37\% \\
Low H\&E color similarity & 2 & 0.24\% [0.05\%, 0.39\%] & -0.20\% \\
High H\&E color similarity & 0 & -0.32\% [-0.62\%, -0.02\%] & -1.87\% \\
\hline
\end{tabular}
\end{table*}

Considering only the high-performing, highly robust models (n=10), predominantly UNI2-h+\allowbreak ABMIL models, alongside the CTransPath+\allowbreak Wagner2023 model and two H-Optimus-1+\allowbreak ABMIL models remained (Figure \ref{fig:fig4-robustness-performance}b). Among these, reference performance was very similar: the top-performing CTransPath+\allowbreak Wagner2023 model achieved an AUC of 0.911, closely followed by UNI2-h+\allowbreak ABMIL models (0.902-0.911), and slightly lower-performing H-Optimus-1+\allowbreak ABMIL models (0.902-0.905). In contrast,  robustness differed more clearly across models: UNI2-h+\allowbreak ABMIL models showed consistently smaller robustness ranges (0.009-0.013) than H-Optimus-1+\allowbreak ABMIL (0.020-0.024) and CTrans\-Path+\allowbreak Wagner2023 (0.021). Bootstrap analysis reinforced this separation: whereas 95\% confidence intervals (CIs) for reference AUC largely overlapped across top-performing models (Figure \ref{fig:fig4-robustness-performance}b, CTransPath+\allowbreak Wagner2023: [0.869,0.950]; best UNI2-h+\allowbreak ABMIL: [0.875,0.944]; best H-Optimus-1+\allowbreak ABMIL: [0.865,0.941]), robustness CIs were narrower and more discriminative. Median robustness CI width was smallest for UNI2-h+\allowbreak ABMIL (0.019), compared with H-Optimus-1+\allowbreak ABMIL (0.031) and CTransPath+\allowbreak Wagner2023 (0.030), indicating greater stability.

\subsection{Impact of staining conditions on model performance}\label{subsec:results-staining-conditions}
Finally, we assessed how each of the four simulated staining conditions influenced model performance compared to the reference. Across all models (n=306), the high intensity condition most frequently yielded the highest AUC (127 models), followed by low H\&E color similarity (51 models), whereas the reference condition was optimal in 65 models (Table \ref{tab:tab2-all-models-staining}). At the same time, each condition was associated with performance decreases in subsets of models, with worst-case scenarios of more than 3\% performance decrease for any condition (Table \ref{tab:tab2-all-models-staining}). Restricting to high-performing, robust models (n=10), high intensity remained the most favorable condition (8/10 models, median $\Delta$AUC +0.50\%, Table \ref{tab:tab2-all-models-staining}, Appendix Figure E.1a), achieving the overall highest AUCs on the SurGen dataset for the CTransPath+Wagner2023 model and a single H-Optimus-1+\allowbreak ABMIL model (both: 0.920), exceeding their respective reference-condition performance (CTransPath+\allowbreak Wagner2023: 0.911, H-Optimus-1+\allowbreak ABMIL: 0.904). Simultaneously, low intensity and high H\&E color similarity led to modest median decreases (-0.46\% and -0.32\%, Table \ref{tab:tab2-all-models-staining}, Appendix Figure E.1b). Worst-case AUC drops also followed a condition-dependent pattern: decreases were minimal (<0.4\%) for high intensity and distinct H\&E color conditions, but larger (>1\%) for low intensity and similar H\&E color conditions.

\section{Discussion}\label{sec:Discussion}
This study provides an evaluation protocol to 1) quantify the robustness of computational pathology models to H\&E staining variability and to 2) systematically assess how staining conditions affect model performance. Across a large population of MSI classification models, we observed substantial heterogeneity in robustness against staining variability, with only a weak inverse relationship between performance and robustness. This suggests that higher predictive performance does not guarantee stability under staining shifts, supporting the notion that robustness to staining variation is a partially independent property of trained models \cite{Lin2025,Stacke2021}. Notably, we found that this pattern persists even in foundation-model-based pipelines, and becomes particularly evident when evaluating many downstream predictors under simulated staining conditions, rather than focusing on a single best-performing model. Unexpectedly, UNI2-h and H-Optimus-1 foundation models were found to be more robust than Virchow2, contrasting prior reports \cite{Jong2025,Koemen2025}. Furthermore, the Wagner2023 model, though built on the less robust CTransPath foundation model \cite{Koemen2025}, ranked among the top three in both performance and robustness to staining variation, suggesting that aggregator training can mitigate foundation model sensitivity to staining variability. Overall, we found that high-performing, robust models exist and can be identified by considering both metrics jointly. 
Systematic trends in model performance emerged across staining conditions. Across all models, high intensity and low H\&E color similarity conditions more frequently yielded top performance, whereas low intensity and high H\&E color similarity tended to produce slightly negative shifts relative to the reference. However, these effects were not universal, as all conditions produced the best AUCs for subsets of models. Considering only high-performing, robust models, the same patterns were more pronounced: high intensity and low H\&E color similarity most often corresponded to the highest AUCs, while low intensity and high H\&E color similarity produced modest decreases. Nonetheless, individual models continued to deviate from these trends, indicating that the effect of staining conditions remains model-dependent and should be assessed individually.
Further, our analysis adds a key piece in understanding the origin of staining variability: in the PLISM dataset hematoxylin intensity was mainly driven by the 13 staining protocols, while stain color similarity and eosin intensity were more strongly influenced by the scanner choice out of seven devices. Since scanners are often fixed, whereas staining protocols can be more easily adjusted, identifying staining conditions that perform reliably with the scanners in use is crucial. 
The observed H\&E staining properties of slides under the same staining protocol may change depending on the scanner, highlighting the need to validate both the staining workflow and its interaction with the lab’s scanners to ensure robust model performance.

Prior work on evaluating CPath model robustness to staining variation has focused on using image-based references \cite{Schoemig-Markiefka2021,Vu2022}, generative \cite{Springenberg2023}, or physically repeated staining protocols \cite{Chai2026,Sun2023,Ruiz2023}. While establishing the clinical relevance of staining variability, they do not anchor perturbations to quantifiable stain intensities or similarity measures within a defined reference space. Consequently, attribution of performance changes to specific staining properties remains challenging. In contrast, our protocol applies deterministic, reference-based staining simulations that explicitly control stain intensity and similarity, enabling systematic testing of trained models under defined staining conditions. This enables interpretable assessment of robustness to staining variation and provides a structured, reference-based evaluation protocol that complements existing training- and evaluation-focused approaches.

Our study has several limitations. First, the PLISM reference library captures only part of the staining variability observed in clinical slides, with the SurGen cohort often exceeding its range for H\&E stain angles, indicating that capturing additional stain preparation protocols and scanner combinations would be beneficial. While larger reference sets with pre-extracted stains exist \cite{Marini2023}, they do not allow verification of color correctness, lack intensity properties as well as traceable diagnostic conditions. PLISM, in contrast, provides known staining properties and scanner origins across a broad range of tissue types. Second, this protocol is focused on staining properties only and does not systematically assess other QC factors such as tissue folds, blur or air bubbles, or scanner-induced effects on sharpness and image quality. However, prior work shows that staining is among the strongest factors influencing model performance \cite{Wang2021,Foote2022,Schoemig-Markiefka2021,Vu2022}. Third, the protocol currently only models H\&E staining properties in standard tissue regions. Other components on the slide, such as blood, necrosis, or heavily pigmented areas, behave differently under staining variation and when acquired with different scanners and would require more sophisticated unmixing methods for accurate modeling. Lastly, only four discrete staining conditions were tested, limiting exploration of continuous or non-linear effects. 

\section{Conclusion}\label{sec:Conclusion}
We present a protocol for systematically quantifying the robustness to H\&E staining variation of computational pathology models, revealing how specific H\&E staining properties impact performance across a broad set of MSI classification models. Our analysis suggests that robustness to H\&E staining variation is a model-specific property, only partially correlating with overall predictive performance, and highlights the interplay between staining protocols and scanner choice in shaping digital appearance. Staining conditions produced trends in model performance: in our experiments, high intensity was associated with slightly better performance, while low intensity or highly similar H\&E colors tended to cause modest decreases. However, this was not universal, with subsets of models showing reverse effects, highlighting the need for model-specific evaluation. %Notably, even the highest-performing models could achieve modest improvements (e.g., +0.9\% AUC) under favorable staining conditions.
These results demonstrate that the proposed protocol can serve as a universal stress-testing tool, revealing both general trends and model-specific sensitivities, and providing actionable guidance for robust deployment of CPath models. 

Our findings highlight actionable directions for both clinical implementers and algorithm developers in advancing stain-robust computational pathology. For clinical implementers, we recommend characterizing the staining properties of routine WSIs, including H\&E intensity and stain vectors, over time. Continuous monitoring via QC protocols can identify operational ranges and outliers, enabling targeted re-staining or re-scanning. Local validation sets can be built from observed slides or by simulating stains at the edges of these operational ranges using protocols like ours, helping ensure consistent laboratory output and robust deployed models within expected staining changes. For algorithm developers, creating reference libraries that capture realistic, traceable, and institution-relevant staining conditions is critical. Public datasets often lack intensity information or include unrealistic color variations, limiting their utility for QC. Combining publicly available references with locally characterized staining properties enables defining operational ranges in which models perform reliably. Ideally, standardized reference sets could be harmonized across institutions and countries to define minimal “fit for purpose” robustness requirements for CPath models, supporting domain generalization, regulatory confidence, and trust in clinical deployment. Future work should extend robustness evaluation beyond rank-based discriminative performance to include absolute score stability, calibration shifts, and scaling drifts under staining changes. Expanding these analyses to other tasks,  modalities (e.g. IHC), and systematically studying interactions between staining, scanning, and training strategies would provide a more complete understanding of model sensitivity and guide both algorithm development and clinical quality assurance, supporting reliable and generalizable deployment of CPath models.

\section*{Acknowledgments}
The results published here are in whole or part based upon data generated by the TCGA Research Network: \url{https://www.cancer.gov/tcga}. We thank all the patients who contributed samples to the SurGen study and TCGA, enabling this research.

\section*{Declaration of competing interest}
L.S. reports travel reimbursement for participation as an invited speaker for Indica Labs. T.B and N.H. received research funding from the Dutch Cancer Society. T.B is supported by the Hanarth Fund, unrelated to the present study. V.K. is a participant in several patent applications on the assessment of cancer biomarkers by digital pathology. T.B., V.K. and N.H. are participants on a patent application on multimodal deep learning for the prediction of recurrence risk in endometrial cancer patients. V.K. reports sponsored research agreements with Roche and IAG, honoraria or consulting fees from Takeda and Roche, and participation as an invited speaker for Sharing Progress in Cancer Care and Indica Labs, all unrelated to the present study. The other authors declare no competing interests.

\section*{Declaration of generative AI and AI-assisted technologies in the manuscript preparation process}
During the preparation of this work the authors used ChatGPT in order to structure and polish the text. After using this tool, the authors reviewed and edited the content as needed and take full responsibility for the content of the published article.

\section*{Funding}
This research did not receive any specific grant from funding agencies in the public, commercial, or not-for-profit sectors.

%\appendix

%\section{}\label{}

\printcredits

\bibliographystyle{cas-model2-names}

\bibliography{cas-refs}



\end{document}